\documentclass[letterpaper]{article} %
\usepackage{aaai26_arxiv}
\usepackage{times}  %
\usepackage{helvet}  %
\usepackage{courier}  %
\usepackage[hyphens]{url}  %
\usepackage{graphicx} %

\usepackage{subfiles} %
\usepackage{makecell} %
\usepackage{multirow} %
\usepackage{subcaption}
\usepackage{color,xcolor}
\usepackage{booktabs}
\usepackage{pifont}

\usepackage{amsmath}
\usepackage{amssymb}
\usepackage{mathtools}
\usepackage{amsthm}

\usepackage{natbib}  %
\usepackage{caption} %
\usepackage{array}
\newcolumntype{C}{>{\centering\arraybackslash}p{1.4cm}}
\usepackage{algorithm}
\usepackage{algorithmic}
\usepackage{adjustbox}

\usepackage{xcolor,colortbl}
\definecolor{citecolor}{HTML}{0071BC}
\definecolor{linkcolor}{HTML}{ED1C24}

\usepackage{newfloat}
\usepackage{listings}
\DeclareCaptionStyle{ruled}{labelfont=normalfont,labelsep=colon,strut=off} %
\floatstyle{ruled}
\newfloat{listing}{tb}{lst}{}
\floatname{listing}{Listing}
\usepackage[pagebackref=false, breaklinks=true, colorlinks, citecolor=citecolor, linkcolor=linkcolor, bookmarks=false]{hyperref}

\title{Pairing-free Group-level Knowledge Distillation for Robust Gastrointestinal Lesion Classification in White-Light Endoscopy}
\author{
    Qiang Hu\equalcontrib,
    Qimei Wang\equalcontrib,
    Yingjie Guo,
    Qiang Li,
    Zhiwei Wang\thanks{Corresponding author. (\url{zwwang@hust.edu.cn})}
}
\affiliations{Wuhan National Laboratory for Optoelectronics, Huazhong University of Science and Technology
}

\begin{document}

\maketitle

\begin{abstract}
White-Light Imaging (WLI) is the standard for endoscopic cancer screening, but Narrow-Band Imaging (NBI) offers superior diagnostic details. A key challenge is transferring knowledge from NBI to enhance WLI-only models, yet existing methods are critically hampered by their reliance on paired NBI-WLI images of the same lesion, a costly and often impractical requirement that leaves vast amounts of clinical data untapped.
In this paper, we break this paradigm by introducing PaGKD, a novel Pairing-free Group-level Knowledge Distillation framework that that enables effective cross-modal learning using unpaired WLI and NBI data. Instead of forcing alignment between individual, often semantically mismatched image instances, PaGKD operates at the \emph{group level} to distill more complete and compatible knowledge across modalities.
Central to PaGKD are two complementary modules: (1) Group-level Prototype Distillation (GKD-Pro) distills compact group representations by extracting modality-invariant semantic prototypes via shared lesion-aware queries; (2) Group-level Dense Distillation (GKD-Den) performs dense cross-modal alignment by guiding group-aware attention with activation-derived relation maps.
Together, these modules enforce global semantic consistency and local structural coherence without requiring image-level correspondence.
Extensive experiments on four clinical datasets demonstrate that PaGKD consistently and significantly outperforms state-of-the-art methods, achieving relative AUC improvements of $3.3\%$, $1.1\%$, $2.8\%$, and $3.2\%$, respectively, establishing a new direction for cross-modal learning from unpaired data.
\end{abstract}

\begin{links}
    \link{Code}{https://github.com/Huster-Hq/PaGKD}
\end{links}

\section{Introduction}
\label{sec:intro}
Endoscopy is a key tool for early detection of gastrointestinal (GI) cancers~\cite{schmitz2022artificial}. Accurate lesion classification from endoscopic images is critical for timely diagnosis and treatment planning; however, manual evaluation is inherently subjective and labor-extensive, driving the development of automated diagnostic systems.
Modern endoscopy relies on two imaging modalities: White-Light Imaging (WLI) and Narrow-Band Imaging (NBI). Among them, NBI enjoys enhanced vascular and mucosal details via spectral filtering, yielding superior lesion visibility and thereby improving classification performance~\cite{tamura2022magnifying,yue2023benchmarking}. However, WLI remains the dominant modality in real-world clinical practice due to its universal availability and default configuration in most endoscopic systems~\cite{soffer2020deep,beg2020diagnosis}, whereas NBI functionality is either absent or underutilized in routine deployments.

This discrepancy raises a pivotal research question: \emph{How can we transfer the rich visual knowledge embedded in NBI to enhance lesion recognition in WLI, particularly when only WLI data is accessible during clinical deployment?} This challenge falls under the broader umbrella of cross-modal knowledge transfer. Emerging studies have initiated investigations into this direction by leveraging NBI as a privileged modality during training to augment WLI-only test-time classification. Representative approaches, such as PolypsAlign~\cite{ma2022toward} and CPC-Trans~\cite{wang2021colorectal}, align modality-specific features using CNN or Transformer architectures guided by lesion priors. More recently, ADD~\cite{hu2025holistic} relaxes the requirement for spatial alignment priors and performs holistic image-level alignment on paired yet unregistered images, achieving notable advancements in WLI-only classification performance.

Despite their efficacy, existing cross-modal distillation methods are limited by their reliance on paired NBI-WLI images of the same lesion. This \emph{instance-specific distillation} paradigm assumes that knowledge can be directly transferred between different modality views of the same physical object. While such paired data is informative, collecting it at scale is costly and impractical. More importantly, most real-world clinical datasets are composed of unpaired NBI and WLI images acquired independently from different lesions and patients, leaving a vast amount of data untapped and severely limiting the scalability of current approaches.

In this work, we address a critical yet underexplored challenge: \emph{cross-instance knowledge distillation} for improving lesion classification in WLI endoscopy. Our goal is to leverage abundant \emph{unpaired} clinical data to transfer NBI-derived knowledge into WLI-only classification model. This is both feasible and clinically meaningful, as many modality-specific patterns in NBI, such as vascular and mucosal structures, exhibit high recurrence across patients and lesions. However, a major bottleneck arises when applying traditional instance-level distillation to unpaired data: individual lesion images often provide only partial and case-specific manifestations of disease. This leads to semantic mismatches and feature incompatibility between the source (NBI) and target (WLI) models, ultimately undermining the effectiveness of cross-modal distillation.

To effectively bridge the modality gap between WLI and NBI images without requiring one-to-one pairing, we propose PaGKD, a \textbf{Pa}iring-free \textbf{G}roup-level \textbf{K}nowledge \textbf{D}istillation framework. Unlike prior methods like ADD~\cite{hu2025holistic}, which align features between individual images, PaGKD operates at the group level by aggregating images with the same lesion class into semantically coherent feature sets. This group-level formulation mitigates sample-level noise and enhances the stability and completeness of cross-modal knowledge transfer.
Specifically, PaGKD is built upon two complementary modules: Group-level Prototype Distillation (GKD-Pro) and Group-level Dense Distillation (GKD-Den). GKD-Pro distills high-level, modality-invariant semantics by extracting class-specific prototypes through shared lesion-aware queries. These prototypes act as global semantic anchors and are aligned across modalities via a contrastive objective, enabling the model to abstract consistent concepts from unpaired groups.
GKD-Den, in contrast, distills local structural knowledge by generating relation maps from refined activation cues, which capture spatial correspondences between modality-specific groups. These maps supervise a group-aware cross-attention mechanism that restructures features across modalities, enforcing alignment at a dense, region-wise level.
Together, GKD-Pro and GKD-Den form a mutually reinforcing mechanism, where GKD-Pro ensures semantic consistency and GKD-Den enforces spatial coherence. This dual-level alignment enables robust knowledge distillation across modalities, even in the absence of image-level pairing.

In summary, our major contributions are as follows.
\begin{itemize}
\item[$\bullet$] We propose PaGKD, a novel pairing-free group-level knowledge distillation framework that enables effective cross-modal learning between WLI and NBI images without requiring one-to-one correspondence. 
\item[$\bullet$] We design two complementary modules: GKD-Pro for modality-invariant semantic prototype learning, and GKD-Den for fine-grained spatial alignment. Together, they achieve coherent, class-consistent feature alignment across WLI and NBI modalities.
\item[$\bullet$] We conduct extensive experiments on three colorectal polyp classification datasets and one gastric cancer dataset. PaGKD consistently outperforms state-of-the-art methods, achieving $3.3\%$, $1.1\%$, $2.8\%$, and $3.2\%$ relative improvements in AUC, respectively.
\end{itemize}

\section{Related Work}
\label{sec:relatedw}
\subsection{Cross-modality Independent Classification (CIC) for GI Diseases}
Significant progress has been made in gastrointestinal (GI) disease classification. Most existing methods treat multi-modal inputs (e.g., WLI and NBI) as modality-agnostic, ignoring potential inter-modal knowledge interaction. Some works focus on architectural improvements, such as integrating multi-level features to enhance classification~\cite{zhang2016automatic}, employing Fourier transforms to mitigate illumination-related noise~\cite{wang2022ffcnet}. Other studies emphasize feature learning via self-supervised strategies: SSL-CPCD~\cite{xu2024ssl} introduces an additive angular margin within cosine similarity to guide patch–image clustering, and SSL-WCE~\cite{guo2020semi} adopts adaptive aggregated attention to extract discriminative features. Additionally, some methods focus on learning local lesion-aware features,such as employing attention mechanisms to integrate local and global representations for classification~\cite{wang2023dlgnet}.
However, these approaches often perform suboptimally on WLI images, primarily due to the lack of effective utilization of the richer diagnostic information in NBI, which limits their classification performance in modality-limited scenarios.

\subsection{Cross-modal Distillation Classification (CDC) for GI Diseases}
Cross-modal knowledge distillation is an effective strategy for transferring knowledge from an informative source modality (NBI) to a less informative target modality (WLI) to improve disease classification. 
Existing CDC methods typically relying on paired images for image-level knowledge transfer. For example, Polypsalign~\cite{wang2021colorectal} adopted discriminator and contrastive loss for global feature alignment; CPC-Trans~\cite{ma2022toward} employed a Transformer-based architecture with cross-attention to align patch-level and global features; and ADD~\cite{hu2025holistic} further performed pixel-wise image-to-image distillation guided by semantic correspondences.
However, NBI functionality is either absent or underutilized in routine deployments, leading to large-scale unpaired data.
These paired-dependent CDC methods exhibit limited generalizability in such settings and fail to fully exploit the unpaired information.
To bridge the modality gap between WLI and NBI without requiring one-to-one pairing, we propose group-level local spatial feature distillation and global prototype distillation, enabling effective and stable unpaired knowledge transfer in challenging GI scenarios.

\section{Methodology}
\label{sec:method}
\begin{figure*}[t]
    \centering
    \includegraphics[width=0.89\linewidth]{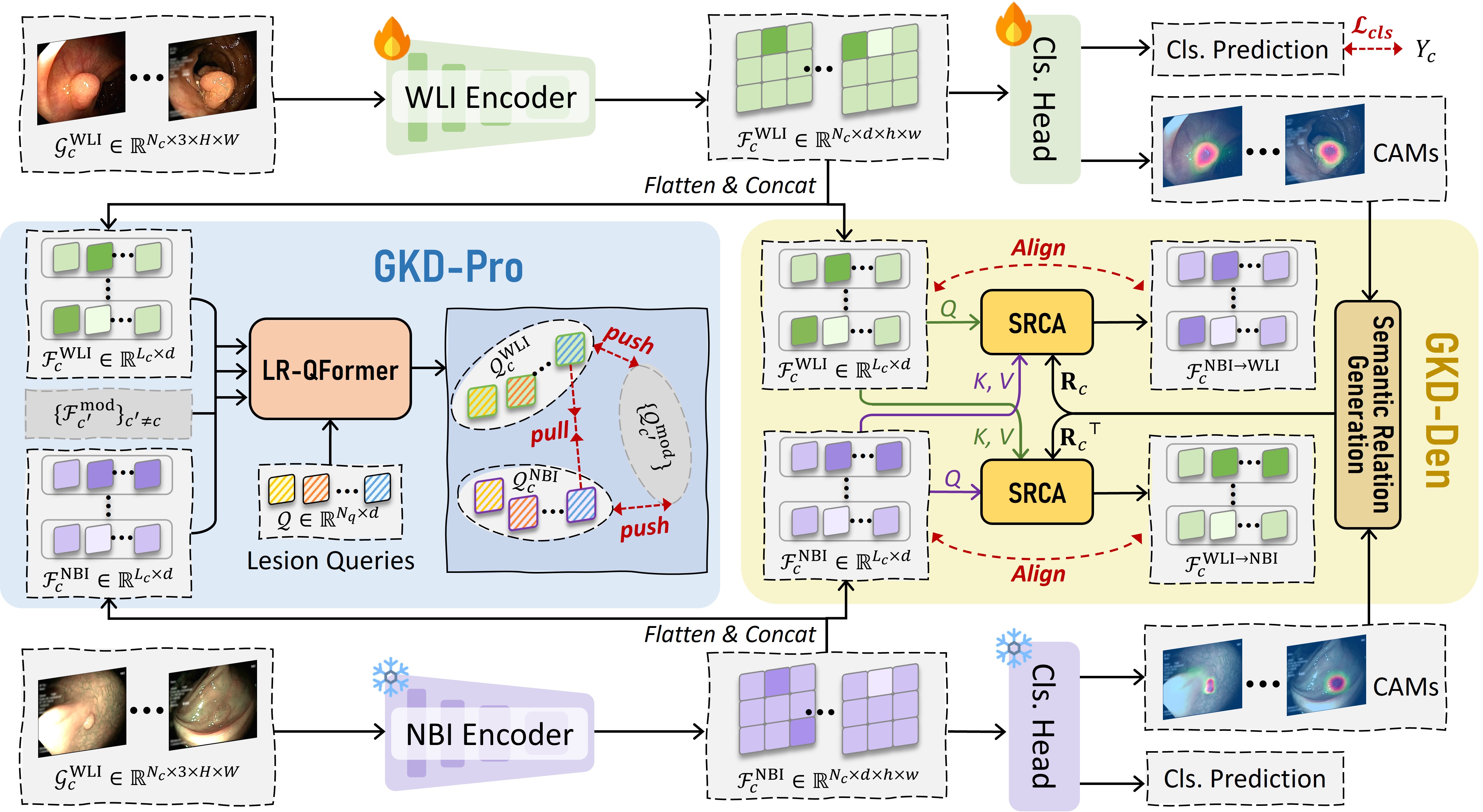}
    \caption{Overview of the proposed PaGKD. It consists of a pre-trained and frozen NBI classifier, a trainable WLI classifier, and two group-level knowledge distillation modules: Group-level Prototype knowledge distillation (GKD-Pro) and Group-level Dense Knowledge Distillation (GKD-Den). During training, given two unpaired image groups from the same class, GKD-Pro collaborates with GKD-Den to achieve robust and multi-granularity cross-modal knowledge distillation between them.}
    \label{fig:method_overview}
\end{figure*}
\subsection{Overview}
As illustrated in Figure~\ref{fig:method_overview}, our framework PaGKD consists of a trainable WLI classifier and a frozen, pretrained NBI classifier. Both classifiers share the same architecture, following ADD~\cite{hu2025holistic}.
To facilitate knowledge transfer between \emph{unpaired} WLI and NBI images, we introduce a group-level feature distillation strategy, instead of regular image-level approaches.
In each training iteration, we construct\footnote{Grouping details are provided in Appendix~\ref{appendix:grouping}.} image groups
$\{\mathcal{G}_c^{\texttt{mod}} \}$,
where $\mathcal{G}_c^{\texttt{mod}}$ contains $N_c$ images from class $c$ in modality $\texttt{mod} \in \{\texttt{WLI}, \texttt{NBI}\}$, and $H$, $W$ denote the input dimensions. Each image within a group is fed independently into its corresponding classifier to extract feature maps: 
\begin{equation}
  \mathcal{G}_c^{\texttt{mod}} \rightarrow \mathcal{F}_{c}^{\texttt{mod}} \in \mathbb{R}^{N_c \times d \times h \times w},
\end{equation}
where $d$, $h$, and $w$ are the channel, height, and width of the feature maps, respectively.

For group-level operations, we flatten and aggregate all spatial features across the group into a unified representation:
\begin{equation}
  \mathcal{F}_c^{\texttt{mod}} = \left\{ \mathbf{f}_{i,c}^{\texttt{mod}} \in \mathbb{R}^{d} \right\}_{i=1}^{L_c}, \quad \text{with } L_c = N_c \cdot h \cdot w.
\end{equation}

We then apply two complementary modules for multi-granularity distillation:
Group-level Prototype Knowledge Distillation (GKD-Pro) performs alignment of global class-level distributions, while Group-level Dense Knowledge Distillation (GKD-Den) aligns local feature.

\subsection{Group-level Prototype Knowledge Distillation}
To facilitate effective cross-modal knowledge transfer from unpaired data, we first group features that share similar lesion traits. While these group-level features offer richer semantics than individual samples, they may still contain redundant or irrelevant patterns. To distill core lesion representations, we extract feature prototypes that summarize the underlying disease characteristics shared across each group.

\subsubsection{Lesion-related Query Transformer for Prototype Representation.}
We propose a learnable prototype extraction mechanism based on the Lesion-Related Query Transformer (LR-QFormer). It uses a set of shared trainable lesion queries $\mathcal{Q}\in \mathbb{R}^{N_q \times d}$ to actively attend to discriminative group features, where $N_q \ll L_c$. Unlike static clustering, this query-based approach enables adaptive and interpretable abstraction of disease semantics.
The queries $\mathcal{Q}$ are shared across all groups and modalities, and are designed to be class- and modality-agnostic. They act as latent anchors representing modality-invariant lesion concepts. During inference, their interaction with modality-specific group features reveals how these lesion attributes are expressed in different imaging conditions.

The LR-QFormer is composed of $T$ stacked Transformer blocks. At each layer, the queries are first refined via self-attention $\texttt{SA}$, then updated by cross-attending $\texttt{CA}$ to the group feature map $\mathcal{F}_c^{\texttt{mod}} \in \mathbb{R}^{L_c \times d}$ with absolute positional encoding $\mathbf{E}_{\texttt{pos}}$~\cite{dosovitskiy2020image}. Formally:
\begin{equation}
\small
\mathcal{Q}_{t,c}^{\texttt{mod}} = \texttt{CA}\left( \texttt{SA}\left(\mathcal{Q}_{t-1,c}^{\texttt{mod}} \right),\ \mathcal{F}_c^{\texttt{mod}} + \mathbf{E}_{\texttt{pos}} \right), ~\mathcal{Q}_{0,c}^{\texttt{mod}} = \mathcal{Q}.
\end{equation}

Through this iterative process, each query accumulates modality-specific evidence of its corresponding disease attribute. After $T$ layers, we obtain the refined set $\mathcal{Q}_{c}^{\texttt{mod}} = \{ \mathbf{q}_{i,c}^{\texttt{mod}} \in \mathbb{R}^{d} \}_{i=1}^{N_q}$ which serves as the prototype representation of the group in modality \texttt{mod}. These prototypes are both compact and interpretable, enabling robust cross-modal alignment in subsequent distillation.

\subsubsection{Group-level Contrastive Loss for Knowledge Distillation.}
To facilitate cross-modal knowledge distillation, we introduce a group-level contrastive objective over the refined lesion query sets $\{\mathcal{Q}_{c}^{\texttt{mod}} \}$ generated for each class $c$ and modality $\texttt{mod} \in \{\texttt{WLI}, \texttt{NBI}\}$. The core motivation is that the same set of lesion queries $\mathcal{Q}$, shared across groups, captures modality-invariant disease semantics. When interacting with group features from different classes and modalities, these queries produce class-specific prototype representations, which reflect how a disease manifests under a particular imaging modality.

Our goal is to align these group-level representations across modalities for the same class, while separating them across different classes. This encourages the WLI classifier to focus on pathology-relevant patterns distilled from NBI, avoiding modality-specific confounders like color or illumination variations.
To this end, we define the similarity between two query sets $\mathcal{Q}_c^{\texttt{mod}}$ and $\mathcal{Q}_{c'}^{\texttt{mod}'}$ as the average cosine similarity between queries at the same index:
\begin{equation}
S_{\mathcal{Q}_{c}^{\texttt{mod}}, \mathcal{Q}_{c'}^{\texttt{mod}'}} = \frac{1}{N_q}\sum\limits_{i = 1}^{N_q} \frac{(\textbf{q}_{i,c}^{\texttt{mod}})^{\top}\textbf{q}_{i,c'}^{\texttt{mod}'}}{\| \textbf{q}_{i,c}^{\texttt{mod}} \| \| \textbf{q}_{i,c'}^{\texttt{mod}'}\|} .
\end{equation}

This formulation assumes each query captures a specific semantic aspect of disease (\textit{e.g.}, boundary irregularity, vascular pattern), enabling a fine-grained alignment across groups.
We then construct a symmetric contrastive loss that encourages high similarity between WLI-NBI pairs of the same class and low similarity between different classes:
\begin{equation}
\small
\begin{split}
    \mathcal{L}_{pro}=-\frac{1}{2C} \sum\limits_{c = 1}^{C} \big( \texttt{log}\frac{\mathrm{exp}({S}_{\mathcal{Q}_{c}^{\texttt{WLI}}, \mathcal{Q}_{c}^{\texttt{NBI}}})}{\sum_{\mathcal{Q}_{c'}^\texttt{mod} \neq \mathcal{Q}_{c}^\texttt{WLI}}\mathrm{exp}({S}_{\mathcal{Q}_{c}^{\texttt{WLI}}, \mathcal{Q}_{c'}^{\texttt{mod}}} )} \\
    + \texttt{log}\frac{\mathrm{exp}({S}_{\mathcal{Q}_{c}^{\texttt{NBI}}, \mathcal{Q}_{c}^{\texttt{WLI}}})}{\sum_{\mathcal{Q}_{c'}^\texttt{mod} \neq \mathcal{Q}_{c}^\texttt{NBI}}\mathrm{exp}({S}_{\mathcal{Q}_{c}^{\texttt{NBI}}, \mathcal{Q}_{c'}^{\texttt{mod}}})}\big).
\end{split}
\end{equation}
This loss term guides the lesion queries to distill disease-centric semantics shared across modalities, thus enabling group-level alignment and effective cross-modal transfer, even under unpaired supervision.

\subsection{Group-level Dense Knowledge Distillation}
While the above-introduced lesion-related queries capture high-level, modality-invariant attributes, they may overlook fine spatial details critical for accurate diagnosis. To complement this, we perform dense alignment between unpaired cross-modal feature groups. In medical imaging, subtle local patterns such as vascular distortion or mucosal texture are often decisive and require pixel-level modeling.

Our approach draws inspiration from ADD~\cite{hu2025holistic}, which aligns images based on Class Activation Maps (CAMs)~\cite{zhou2016learning}. While ADD operates on individual images, we extend this idea to the group level by aligning all WLI and NBI features within the same class. This enables more stable and comprehensive cross-modal supervision.

\subsubsection{Semantic Relation Generation for Inter-class Dense Connection.}
For each group $\mathcal{G}_c^{\texttt{mod}}$ in class $c$, we first compute CAMs from the extracted features, then apply pixel-adaptive refinement~\cite{ru2022learning} to enhance spatial coherence. The refined CAMs are flattened into a sequence $\mathcal{M}_c^{\texttt{mod}} = \{ \mathbf{m}_{i,c}^{\texttt{mod}} \}_{i=1}^{L_c}$, and discretized using a tri-thresholding scheme with thresholds $\tau_1$ and $\tau_2$. Each element $\mathbf{m}_{i,c}^{\texttt{mod}} \in \{0,1,\varnothing\}$ indicates background, lesion, or ambiguity.

To construct a semantic relation matrix $\mathbf{R}_c \in \{0,1\}^{L_c \times L_c}$ between WLI and NBI groups, we mark a pair of positions $(p,q)$ as semantically aligned (\textit{i.e.}, $r_{(p,q),c}=1$) if both are confidently labeled (non-ambiguous) and share the same semantic label in their respective binarized CAMs:
\begin{equation}
\small
    \mathbf{R}_{c}(p,q) = \begin{cases}
  0& \text{ if }~\mathbf{m}_{p,c}^{\texttt{WLI}} = \mathbf{m}_{q,c}^{\texttt{NBI}} ~\text{and}~\mathbf{m}_{p,c}^{\texttt{WLI}},\mathbf{m}_{q,c}^{\texttt{NBI}} \ne \varnothing,\\
  -\infty& \text{ otherwise }.
  \end{cases}
\end{equation}

This relation matrix serves as dense supervision for pixel-level cross-modal distillation. It enables the model to transfer lesion-related knowledge from NBI to WLI at a finer granularity, even in the presence of spatial misalignment or modality-specific artifacts.

\subsubsection{Semantic Relation-guided Cross-Attention (SRCA) for Dense Distillation.}
Given the semantic relation matrix $\mathbf{R}_c$, we introduce a Semantic Relation-guided Cross-Attention (SRCA) module to align dense local features across unpaired cross-modal groups. This module uses relation-aware attention to reshape spatial structures and enhance semantic consistency during feature interaction.
Formally, given group features $\mathcal{F}_c^{\texttt{WLI}}$ and $\mathcal{F}_c^{\texttt{NBI}}$, we reconstruct a spatially `aligned' version of NBI in the WLI coordinate space via:
 \begin{equation}
 \small
    \begin{aligned}
        &\mathcal{F}_c^{\texttt{NBI} \xrightarrow{}\texttt{WLI}}=\mathbf{A}(\mathcal{F}_c^{\texttt{NBI}}W_v) \in \mathbb{R}^{ L_{c} \times d}, \\
        &\mathbf{A} = \texttt{Softmax}\bigg(\mathbf{R}_c + \frac{{(\mathcal{F}_c^{\texttt{WLI}}W_q)}(\mathcal{F}_c^{\texttt{NBI}}W_k)^{\top}}{\sqrt{{d}/4}}\bigg),
    \end{aligned}
\label{eq:2}
\end{equation}
where $W_q, W_k \in \mathbb{R}^{d \times \frac{d}{4}}$ and $W_v \in \mathbb{R}^{d \times d}$ are linear projections. The attention matrix $\mathbf{A} \in \mathbb{R}^{L_c \times L_c}$ guides the spatial reconstruction by aggregating semantically matched positions from NBI features, reshaping them to better align with WLI semantics. 
A symmetric operation is performed to reconstruct WLI features in the NBI space, \textit{i.e.}, $\mathcal{F}_c^{\texttt{WLI} \rightarrow \texttt{NBI}}$, using the transposed relation matrix $\mathbf{R}_c^\top$. This bidirectional mapping supports dense cross-modal feature fusion under weakly supervised and unpaired conditions.

Compared with ADD~\cite{hu2025holistic}, which aligns single images using CAM-derived correspondences, our method performs attention-guided alignment at the group level, allowing the model to query from richer spatial and semantic contexts. Importantly, our spatial reshaping aggregates evidence across groups, enabling more reliable reconstruction even when individual samples lack complete lesion information.
With the two sets of group-level `reconstructed' features $\mathcal{F}_c^{\texttt{NBI}\rightarrow \texttt{WLI}} = \{ \mathbf{f}_{i,c}^{\texttt{NBI}\rightarrow \texttt{WLI}} \}_{i=1}^{L_c}$ and $\mathcal{F}_c^{\texttt{WLI}\rightarrow \texttt{NBI}} = \{ \mathbf{f}_{i,c}^{\texttt{WLI}\rightarrow \texttt{NBI}} \}_{i=1}^{L_c}$ obtained, we define a bidirectional consistency loss to supervise this alignment as formulated in Eq.~(\ref{eq:8}).
This loss supervises dense distillation at the group level in both directions and encourages fine-grained cross-modal consistency.
\begin{equation}
\small
\begin{split}
\mathcal{L}_{den} = \frac{1}{C (L_c)^2 }\sum\limits_{c = 1}^{C} \bigg( 
\sum\limits_{p = 1}^{L_{c}} \left\| \mathbf{f}_{p,c}^{\texttt{WLI}} - \mathbf{f}_{p,c}^{\texttt{NBI}\rightarrow \texttt{WLI}} \right\|_2 \\
+ \sum\limits_{q = 1}^{L_{c}} \left\| \mathbf{f}_{q,c}^{\texttt{NBI}} - \mathbf{f}_{q,c}^{\texttt{WLI}\rightarrow \texttt{NBI}} \right\|_2 
\bigg).
\end{split}
\label{eq:8}
\end{equation}

\subsection{Network Training and Inference}
In the training stage, in addition to GKD-Pro and GKD-Den distillations, \textit{i.e.}, $\mathcal{L}_{pro}$ and $\mathcal{L}_{den}$, our overall training objective also includes a cross-entropy loss term $\mathcal{L}_{cls}$ for the classification results of the WLI classifier for the ground truth class labels.
The total training loss $\mathcal{L}_{total}$ is formulated as:
\begin{equation}
\small
    \mathcal{L}_{total} = \mathcal{L}_{pro} +\mathcal{L}_{den} + \mathcal{L}_{cls},
\end{equation}
where $\mathcal{L}_{cls} = \frac{1}{C}\sum_{c=1}^{C}\big(\frac{1}{N_{c}}\sum_{i=1}^{N_c} \texttt{CE}(P_{i,c}^{\texttt{WLI}}, Y_c)\big)$.
Here, $\texttt{CE}(\cdot,\cdot)$ is the cross-entropy loss function, $P_{i,c}^{\texttt{WLI}}$ is the classification result of the WLI classifier for the $i$-th image in the WLI image group of class $c$, and $Y_c$ is the one-hot label corresponding to the class $c$.
In the inference stage, we only keep the WLI classifier, allowing it to receive a single WLI image and output the classification result.

\section{Experiment}
\label{sec:exp}
\subsection{Datasets and Evaluation Metrics}
\begin{table}
\begin{center}
\footnotesize
\setlength{\tabcolsep}{3pt}
\begin{tabular}{@{}l l l ccc@{}}
\hline

\hline
\multirow{2}{*}{{Type}} & \multirow{2}{*}{{Dataset}} & \multirow{2}{*}{{Class}} & \multirow{2}{*}{\makecell[c]{WLI-NBI\\Pairs}} & \multicolumn{2}{c}{{Unpaired}} \\
 & & & & {WLI} & {NBI} \\
\hline
\multirow{7}{*}{\rotatebox{90}{Colorectal Polyp}} 
 & \multirow{2}{*}{PolypSet} 
   & Hyperplasia & $63$ & $200$ & $200$ \\
 & & Adenoma  & $102$  & $250$ & $250$ \\
\cline{2-6}
 & \multirow{3}{*}{PICCOLO} 
   & Hyperplasia & $249$ & $155$ & $35$ \\
 & & Adenoma       & $620$  & $834$ & $162$ \\
 & & Adenocarcinoma & $186$  & $76$ & $17$ \\
\cline{2-6}
 & \multirow{2}{*}{IH-Polyp} 
   & Hyperplasia & $257$ & $1,250$ & $427$ \\
 & & Adenoma  & $299$  & $2,480$ & $494$ \\
\hline
\multirow{2}{*}{\rotatebox{90}{GC}} 
 & \multirow{2}{*}{IH-GC} 
   & Gastric Inflammation & $120$ & $215$ & $159$ \\
 & & Gastric Cancer   & $144$  & $254$ & $144$ \\
\hline

\hline
\end{tabular}
\caption{Statistics of experimental datasets.}
\label{table:dataset}
\end{center}
\end{table}

\subsubsection{Datasets.}
We conduct experiments on four gastrointestinal (GI) cancer datasets independently, including three colorectal polyp datasets: PICCOLO~\cite{sanchez2020piccolo}, PolypSet~\cite{li2021colonoscopy}, and  an in-house dataset, denoted IH-Polyp, and an in-house gastric cancer (GC) dataset, denoted IH-GC.
Detailed statistics are summarized in Table~\ref{table:dataset}.

\subsubsection{Evaluation Metrics.}
To fairly compare the performance of the NBI and WLI classifiers, we perform $5$-fold cross-validation on the paired data in all datasets, while the unpaired data are only used as the train set.
We ensure that the patient-level separation between the train and test sets.
To evaluate performance, we use four metrics: Accuracy (Acc), Precision (Pre), Recall (Rec), F$1$-score (F$1$), and Area Under the Curve (AUC).
Unless otherwise specified, all evaluations are conducted on the WLI test set.

\subsection{Implementation Details}
PaGKD is implemented using PyTorch~\cite{paszke2019pytorch} and trained on a single NVIDIA $4090$ GPU with $24$GB memory.
Weights of ResNet-50~\cite{he2016deep} pre-trained on ImageNet~\cite{deng2009imagenet} are loaded as initialization of the WLI classifier.
NBI classifier is pretrained on NBI train set and frozen during distillation.
The input size is set to $448\times448$, and the batch size is set to $24$.
We define the group size for each class within a batch based on the class distribution in the dataset.
The number of epochs is set to $100$ and the initial learning rate is set to $1e^{-4}$.
The optimizer is set to Adam~\cite{kingma2014adam} with $1e^{-8}$ weight decay.
We set the number of lesion queries in LR-QFormer $N$ as $12$, and set $\tau_1$ and $\tau_2$ as $0.3$ and $0.7$, respectively, according to the parameter search experiments.
Note that PaGKD can also be seamlessly extended to paired data and leverage the inherent prior pairing relationships by simply transforming the input features of GKD-Pro and GKD-Den from group-level to image-level.

\subsection{Comparison with State-of-the-arts (SOTAs)}
\begin{table*}[t]
\begin{center}
\renewcommand\arraystretch{1}
\setlength{\tabcolsep}{0.8 pt}
\footnotesize
\begin{tabular}{ll|ccccc|ccccc|ccccc|ccccc}
\hline

\hline
\multirow{2}{*}{Methods} &\multirow{2}{*}{Train Data} &\multicolumn{5}{c|}{PICCOLO ($3$~Classes)} &\multicolumn{5}{c|}{PolypSet ($2$~Classes)} & \multicolumn{5}{c|}{IH-Polyp ($2$~Classes)} & \multicolumn{5}{c}{IH-GC ($2$~Classes)}\\
& &Acc &Pre &Rec &F$1$ &AUC &Acc &Pre &Rec &F$1$ &AUC &Acc &Rec &Pre &F$1$ &AUC &Acc &Rec &Pre &F$1$ &AUC\\
\hline
NBI Classifier &$ $ &$80.7$ &$79.1$ &$79.6$ &$79.1$ &$86.9$ &$94.8$ &$96.0$ &$95.1$   &$95.5$ &$97.6$ &$83.8$ &$84.3$ &$89.4$  &$86.8$ &$87.0$ &$81.9$ &$76.7$ &$89.7$&$83.5$&$86.3$\\
\hline
\multicolumn{8}{l}{$\blacktriangleright$~\emph{Cross-modal Independent Classification (CIC) methods:}}\\
SSL-CPCD &$\mathcal{D}_{p}+\mathcal{D}_{unp}$ &$64.1$ &$65.4$ &$51.4$ &$54.0$ &$72.5$ &$76.7$ &$77.7$ &$85.3$  &$81.3$ &$80.2$ &$64.9$ &$70.1$ &$71.2$ &$70.6$ &$68.9$ &$66.7$ &$74.1$ &$75.5$ &$74.8$ &$70.4$\\
SSL-WCE &$\mathcal{D}_{p}+\mathcal{D}_{unp}$ &$66.0$ &$67.8$ &$57.5$ &$59.4$ &$73.9$ &$76.2$ &$76.5$ &$86.3$ &$81.1$ &$81.2$ &$56.8$ &$57.1$ &$90.3$ &$70.0$ &$57.5$ &$64.9$ &$73.9$ &$70.8$ &$72.3$ &$69.6$\\
FFCNet &$\mathcal{D}_{p}+\mathcal{D}_{unp}$ &$76.8$ &$80.2$ &$68.3$ &$72.1$ &$80.1$ &$87.2$ &$90.8$ &$87.3$ &$89.0$ &$91.4$ &$64.0$ &$66.7$ &$78.8$ &$72.2$ &$62.9$  &$70.4$ &$\underline{89.2}$ &$62.3$ &$73.4$ &$70.8$\\
DLGNet$^\dagger$ &$\mathcal{D}_{p}+\mathcal{D}_{unp}$ &$68.2$ &$67.8$ &$57.5$ &$59.4$ &$74.5$ &$80.8$ &$80.5$ &$89.2$ &$84.6$ &$83.4$ &$67.6$ &$74.2$ &$69.7$ &$71.9$ &$63.3$ &$70.4$ &$78.4$ &$75.5$ &$76.9$ &$66.0$\\
\hline
\multicolumn{8}{l}{$\blacktriangleright$~\emph{Cross-modal Distillation Classification (CDC) methods:}}\\
CPC-Trans$^\dagger$ &$\mathcal{D}_{p}$ &$78.6$ &$76.7$ &$76.2$ &$76.4$ &$86.6$ &$90.1$ &$89.0$ &$95.1$ &$91.9$ &$87.2$ &$69.4$ &$67.4$ &$\mathbf{93.9}$ &$78.5$ &$75.6$ &$75.6$ &$73.2$ &$\mathbf{90.6}$ &$79.4$ &$75.5$\\
\rowcolor{gray!20}
CPC-Trans$^\dagger$ &$\mathcal{D}_{p}+\mathcal{D}_{unp}$ &$78.2$ &$76.7$ &$76.5$ &$76.6$ &$\underline{87.2}$ &$89.9$ &$88.3$ &$94.7$ &$91.0$ &$86.5$ &$62.2$ &$72.2$ &$59.1$ &$65.0$ &$69.6$ &$67.9$ &$73.3$ &$68.8$ &$71.0$ &$69.8$\\
SAMD &$\mathcal{D}_{p}$ &$75.3$ &$77.0$ &$70.0$ &$72.8$ &$83.1$ &$89.0$ &$89.5$ &$92.2$ &$90.8$ &$90.0$ &$70.3$ &$70.0$ &$74.6$ &$75.8$ &$75.2$ &$73.6$ &$76.7$ &$71.9$ &$74.2$ &$76.6$\\
\rowcolor{gray!20}
SAMD &$\mathcal{D}_{p}+\mathcal{D}_{unp}$ &$74.3$ &$74.7$ &$67.9$ &$70.1$ &$79.3$ &$87.2$ &$89.2$ &$89.2$ &$89.2$ &$90.1$ &$62.2$ &$76.1$ &$53.0$ &$62.5$ &$70.8$ &$69.6$ &$\mathbf{89.5}$ &$53.1$ &$66.7$ &$77.3$\\
PolypsAlign$^\dagger$ &$\mathcal{D}_{p}$ &$79.4$ &$80.4$ &$74.1$ &$76.3$ &$84.7$ &$91.3$ &$93.1$ &$92.2$ &$92.6$ &$90.6$ &$70.3$ &$71.4$ &$83.3$ &$76.9$ &$74.3$ &$73.2$ &$79.3$ &$71.9$ &$75.4$ &$81.4$\\
\rowcolor{gray!20}
PolypsAlign$^\dagger$ &$\mathcal{D}_{p}+\mathcal{D}_{unp}$ &$76.2$ &$\mathbf{88.8}$ &$58.8$ &$63.9$ &$81.5$ &$86.0$ &$91.5$ &$84.3$ &$87.8$ &$85.7$ &$60.4$ &$67.2$ &$65.2$ &$66.2$ &$60.6$ &$71.4$ &$75.0$ &$75.0$ &$75.0$ &$73.4$\\
ADD &$\mathcal{D}_{p}$ &$78.1$ &$79.1$ &$74.4$ &$76.4$ &$83.8$ &$93.4$ &$\mathbf{96.1}$ &$92.8$ &$94.0$ &$93.7$ &$80.2$ &$76.8$ &$\underline{90.5}$ &$\underline{84.1}$ &$\underline{82.8}$ &$75.0$ &$82.1$ &$71.9$ &$79.2$ &$77.3$\\
\rowcolor{gray!20}
ADD &$\mathcal{D}_{p}+\mathcal{D}_{unp}$ &$76.8$ &$77.9$ &$71.7$ &$74.2$ &$84.2$ &$91.9$ &$90.7$ &$96.1$ &$93.3$ &$93.6$ &$74.8$ &$78.8$ &$78.8$ &$78.8$ &$75.7$ &$75.0$ &$76.5$ &$81.2$ &$78.8$ &$77.8$\\
PaGKD (Ours) &$\mathcal{D}_{unp}$ &$\underline{80.8}$ &$80.8$ &$\underline{77.2}$ &$\underline{78.8}$ &$86.6$ &$\underline{94.2}$ &$91.8$ &$\mathbf{99.0}$ &$\underline{95.3}$ &$94.0$ &$79.3$ &$77.9$ &$83.9$ &$83.9$ &$82.4$ &$\underline{79.4}$ &$82.5$ &$\underline{88.0}$ &$\underline{80.5}$ &$\underline{82.2}$\\
PaGKD (Ours) &$\mathcal{D}_{p}$ &$78.6$ &$79.2$ &$73.3$ &$75.3$ &$84.4$ &$93.6$ &$93.3$ &$96.1$ &$94.7$ &$\underline{94.2}$ &$\underline{81.1}$ &$\underline{83.6}$ &$83.0$ &$83.2$ &$81.1$ &$76.8$ &$82.8$ &$75.0$ &$78.7$ &$81.0$\\
PaGKD (Ours) &$\mathcal{D}_{p}+\mathcal{D}_{unp}$ &$\mathbf{81.9}$ &$\underline{81.9}$ &$\mathbf{80.6}$ &$\mathbf{81.1}$ &$\mathbf{90.1}$ &$\mathbf{94.8}$ &$\underline{94.3}$ &$\underline{97.1}$ &$\mathbf{95.7}$ &$\mathbf{94.7}$ &$\mathbf{82.0}$ &$\mathbf{84.8}$ &$84.8$ &$\mathbf{84.8}$ &$\mathbf{85.1}$ &$\mathbf{81.5}$ &$86.5$ &$84.9$ &$\mathbf{85.7}$ &$\mathbf{84.0}$\\
\hline

\hline
\end{tabular}
\caption{Comparisons on three polyp classicication datasets: PICCOLO, PolypSet, and IH-Polyp, and a gastric cancer dataset IH-GC. `NBI Classifier' denotes the performance of the pre-trained NBI classifier on the NBI test set. The gray lines indicate that the methods originally designed for paired data receive additional unpaired train data. ‘$^\dagger$’ represents that the method requires additional prior information about the location of polyps during training and inference. \textbf{Bold} and \underline{underlined} values indicate the best and second-best results, respectively.}
\label{table:compare_SOTAs}
\end{center}
\end{table*}
We compare our PaGKD with eight SOTA methods on four GI cancer datasets, including four cross-modal independent classification (CIC) methods: SSL-CPCD~\cite{xu2024ssl}, SSL-WCE~\cite{guo2020semi}, FFCNet~\cite{wang2022ffcnet}, and DLGNet~\cite{wang2023dlgnet}, and four cross-modal distillation classification (CDC) methods: CPC-Trans~\cite{ma2022toward}, SAMD~\cite{shen2023auxiliary}, PolypsAlign~\cite{wang2021colorectal}, and ADD~\cite{hu2025holistic}.
For fairness, all methods adopt ResNet-50~\cite{he2016deep} as backbone, except CPC-Trans using ViT-S~\cite{dosovitskiy2020image}, which relies on the Transformer architecture.
Since all CIC methods supervise each image independently without relying on paired data, they are trained with paired and unpaired data to maximize the train data.
The CDC methods are inherently designed for paired data.
For comprehensive comparison, we train them in two settings: only paired data and a combination of paired and unpaired data.

The results are presented in Table~\ref{table:compare_SOTAs}, from which we derive three key observations:
(1) The CDC methods outperform CIC methods in terms of overall trends on four datasets, which proves the efficacy of cross-modal knowledge distillation from the NBI modality to the WLI modality in enhancing WLI GI cancer classification.
(2) The introduction of additional unpaired data not only fails to enhance the performance of existing CDC methods but also leads to performance degradation.
This confirms that applying these methods, which are originally designed for paired images of the same lesion instance, to unpaired images from different lesions will introduce noise and hinder effective feature learning.
(3) When trained on the combination of paired and unpaired data, PaGKD significantly outperforms other SOTAs across all four datasets, achieving relative AUC improvements of at least $3.3\%$, $1.1\%$, $2.8\%$, and $3.2\%$, respectively, and achieves performance closest to the NBI classifier.
Furthermore, even when using only unpaired data, PaGKD achieves performance that is competitive with or even better than the best competing methods.
These results demonstrate the effectiveness of PaGKD in utilizing unpaired data.

\subsection{Ablation Studies}
\subsubsection{Effectiveness of Key Components.}
\begin{table}[t]
\begin{center}
\renewcommand\arraystretch{1}
\setlength{\tabcolsep}{3.2 pt}
\footnotesize
\begin{tabular}{cc|ccc|ccc}
\hline

\hline
\multicolumn{2}{c|}{Components}&\multicolumn{3}{c|}{PICCOLO} &\multicolumn{3}{c}{IH-GC}\\
GKD-Pro &GKD-Den &Acc &F$1$ &AUC &Acc &F$1$ &AUC\\
\hline
\ding{55} &\ding{55} &$62.8$ &$52.4$ &$71.2$ &$62.5$ &$61.8$ &$66.9$ \\
\checkmark &\ding{55} &$77.7$ &$75.3$ &$83.5$ &$75.0$ &$80.5$ &$75.5$\\
\ding{55} &\checkmark &$77.3$ &$75.2$ &$85.0$ &$75.0$ &$75.0$ &$77.3$\\
\checkmark &\checkmark &$\mathbf{81.9}$ &$\mathbf{81.1}$ &$\mathbf{90.1}$ &$\mathbf{81.5}$ &$\mathbf{85.7}$ &$\mathbf{84.0}$\\
\hline

\hline
\end{tabular}
\caption{Ablation study on the effectiveness of the two key components: GKD-Pro and GKD-Den.}
\label{table:ablation_key_component}
\end{center}
\end{table}
To verify the effectiveness of the two key components of PAGKD: Group-level Prototype Knowledge Distillation (GKD-Pro) and Group-level Dense Knowledge Distillation (GKD-Den), we train three variants of PaGKD by disabling GKD-Pro or/and GKD-Den on the combination of paired and unpaired data of both PICCOLO and IH-GC.
When both GKD-Pro and GKD-Den are disabled, we actually directly train a ResNet-$50$ classifier in a cross-modal independent manner.
As shown in Table~\ref{table:ablation_key_component}, introducing GKD-Pro or GKD-Den individually produces notable performance improvements, demonstrating the effectiveness of each component in isolation.
When both components are employed jointly, the model achieves further performance gains.
This suggests that their contributions are complementary rather than redundant, enabling effective cross-modal knowledge distillation at both global-level and local-level.

\subsubsection{Effectiveness of Key Sub-components.}
\begin{table}[t]
\begin{center}
\renewcommand\arraystretch{1}
\setlength{\tabcolsep}{4.3 pt}
\footnotesize
\begin{tabular}{l|ccc|ccc}
\hline

\hline
\multirow{2}{*}{Methods} &\multicolumn{3}{c|}{PICCOLO} &\multicolumn{3}{c}{IH-GC}\\
&ACC &F$1$ &AUC &ACC &F$1$ &AUC\\
\hline
{PaGKD} &$\mathbf{81.9}$ &$\mathbf{81.1}$ &$\mathbf{90.1}$ &$\mathbf{81.5}$ &$\mathbf{85.7}$ &$\mathbf{84.0}$\\
\hline
\multicolumn{6}{l}{$\blacktriangleright$~\textit{GKD-Pro:}}\\
\textit{w/o} LR-QFormer &$77.0$ &$74.8$ &$85.7$ &$76.8$ &$76.3$ &$78.3$\\
\hline
\multicolumn{6}{l}{$\blacktriangleright$~\textit{GKD-Den:}}\\
\textit{w/o} SRCA &$76.6$ &$74.7$ &$83.2$ &$75.6$ &$80.0$ &$76.8$\\
\textit{w/o} Bidirectional &$78.8$ &$78.0$ &$87.8$ &$79.4$ &$83.8$ &$81.9$\\
\hline

\hline
\end{tabular}
\caption{Ablation study on the effectiveness of a key sub-component in GKD-Pro: LR-QFormer, and two key sub-components in GKD-Den: SRCA and bidirectional consistency loss (denoted as `Bidirectional').}
\label{table:ablation_sub_component}
\end{center}
\end{table}
To evaluate the effectiveness of the key sub-components in GKD-Pro and GKD-Den, including LR-QFormer, SRCA and bidirectional consistency loss, we conduct ablation studies by individually disabling each component.
Specifically, we ablate LR-QFormer by directly applying the group-level contrastive loss between image feature vectors obtained via average pooling, instead of between prototypes extracted by LR-QFormer.
For SRCA, we replace it with the standard cross-attention mechanism~\cite{vaswani2017attention} without leveraging the semantic relation map.
To assess the effectiveness of the bidirectional consistency loss, we retain only the consistency loss term between $\mathbf{f}_{p,c}^{\mathrm{WLI}}$ and $\mathbf{f}_{p,c}^{\mathrm{NBI} \rightarrow \mathrm{WLI}}$ in Eq.~(\ref{eq:8}).
As shown in Table~\ref{table:ablation_sub_component}, for GKD-Pro, removing LR-QFormer leads to a performance degradation.
This is attributed to the fact that directly pooled features lack sufficient pathology-specific information, thereby weakening the effectiveness of contrastive learning for the model's feature learning in lesion classification.
For GKD-Den, ablating either SRCA or the bidirectional consistency loss results in varying degrees of performance degradation.
The is because that removing SRCA compromises semantic consistency in the reconstruction of the spatially aligned feature, which undermines the reliability of the reconstructed features and weakens the subsequent dense distillation process.
Similarly, using only the unidirectional consistency loss limits the completeness of knowledge transfer from NBI domain to WLI domain, thus reducing the effectiveness of dense knowledge distillation.

\subsubsection{Group-level \textit{vs.} Image-level Knowledge Distillation.}
\begin{table}[t]
\begin{center}
\renewcommand\arraystretch{1}
\setlength{\tabcolsep}{5.7 pt}
\footnotesize
\begin{tabular}{l|ccc|ccc}
\hline

\hline
\multirow{2}{*}{Level} &\multicolumn{3}{c|}{PICCOLO} &\multicolumn{3}{c}{IH-GC}\\
&ACC &F$1$ &AUC &ACC &F$1$ &AUC\\
\hline
\multicolumn{6}{l}{$\blacktriangleright$~\textit{GKD-Pro:}}\\
Group-level &$77.7$ &$75.3$ &$83.5$ &$75.0$ &$80.5$ &$75.5$\\
Image-level &$64.0$ &$64.9$ &$78.6$ &$67.9$ &$71.9$ &$69.0$\\
\hline
\multicolumn{6}{l}{$\blacktriangleright$~\textit{GKD-Den:}}\\
Group-level &$77.3$ &$75.2$ &$85.0$ &$75.0$ &$75.0$ &$77.3$\\
Image-level &$73.9$ &$69.7$ &$79.9$ &$69.6$ &$75.4$ &$71.9$\\
\hline
\multicolumn{6}{l}{$\blacktriangleright$~\textit{Jointly (GKD-Pro$+$GKD-Den):}}\\
Group-level &$\mathbf{81.9}$ &$\mathbf{81.1}$ &$\mathbf{90.1}$ &$\mathbf{81.5}$ &$\mathbf{85.7}$ &$\mathbf{84.0}$\\
Image-level &$75.3$ &$72.1$ &$84.3$ &$73.2$ &$75.4$ &$75.7$\\
\hline

\hline
\end{tabular}
\caption{Comparisons between our proposed group-level distillation components and their image-level variants.}
\label{table:ablation_group_vs_image}
\end{center}
\end{table}
To verify the core idea of PaGKD, \textit{i.e.}, group-level knowledge distillation, we modify GKD-Pro and GKD-Den into image-level variants by replacing the original group inputs with randomly sampled NBI-WLI pairs,
and evaluate their performance both individually and jointly.
The quantitative results are shown in Table~\ref{table:ablation_group_vs_image}. Image-level distillation leads to significant performance drops for both modules when used individually.
Specifically, it hinders GKD-Pro from extracting comprehensive pathological-relevant features via LR-QFormer, and reduces key-value diversity in SRCA, impairing the construction of spatially aligned feature in GKD-Den.
More notably, their joint application under image-level causes more performance degradation.
\begin{figure}[t]
    \centering
    \includegraphics[width=0.75\linewidth]{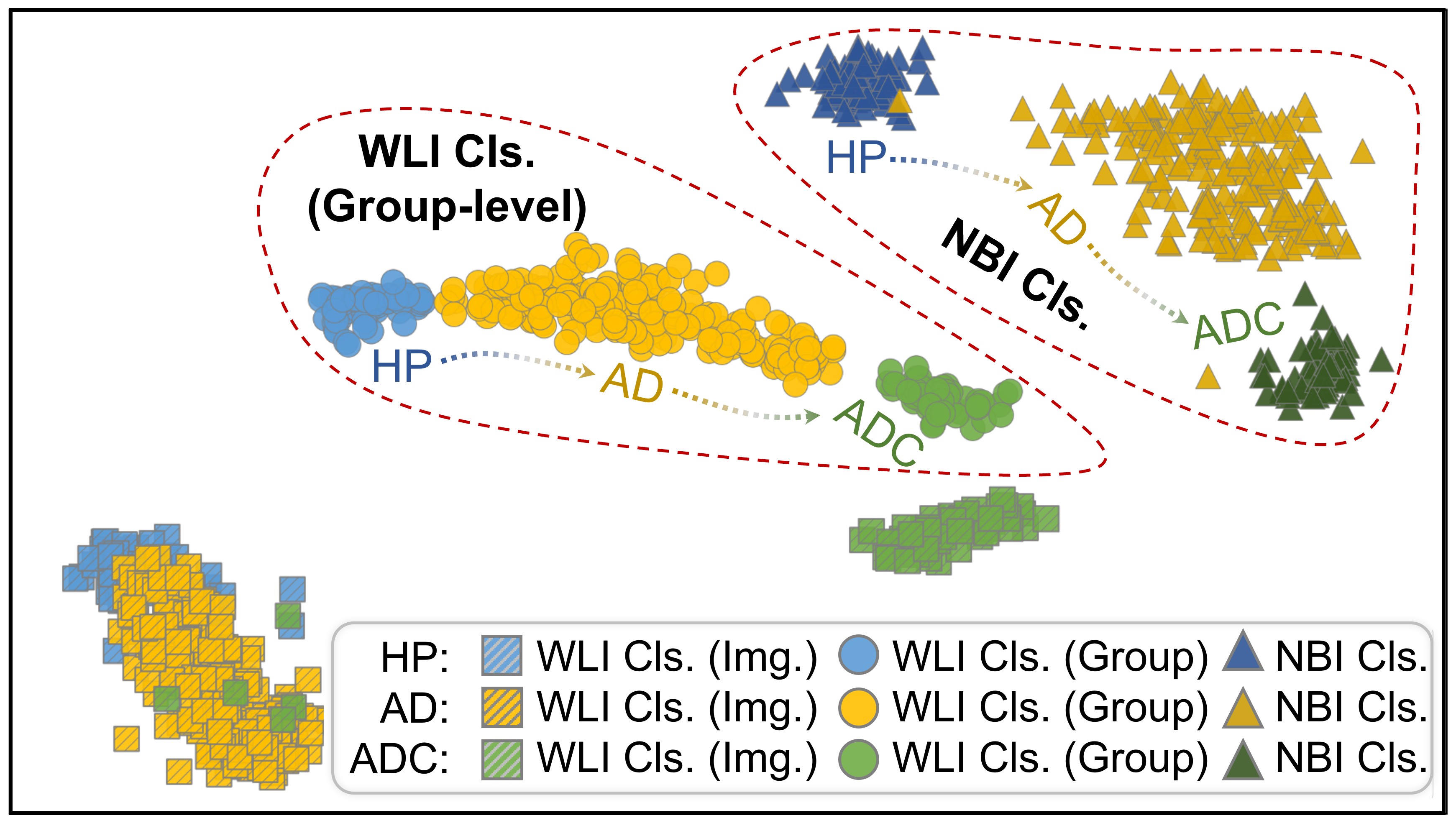}
    \caption{t-SNE visualization of the features on PICCOLO. HP (Hyperplasia) $\rightarrow$ AD (Adenoma) $\rightarrow$ ADC (Adenocarcinoma): the severity of the disease increases progressively.}
    \label{fig:tsne}
\end{figure}
To clearly show the effect of group-level and image-level distillation on the WLI classifier, we visualize the feature distributions using t-SNE~\cite{maaten2008visualizing}. As shown in Figure.~\ref{fig:tsne}, group-level distillation better preserves inter-class separability and spatial structure aligned with the pre-trained NBI features, whereas image-level distillation yields entangled and inconsistent distributions, demonstrating the superiority of group-level distillation under unpaired settings.

\subsubsection{Hyperparameters Configuration.}
\begin{figure}
    \centering
    \includegraphics[width=0.97\linewidth]{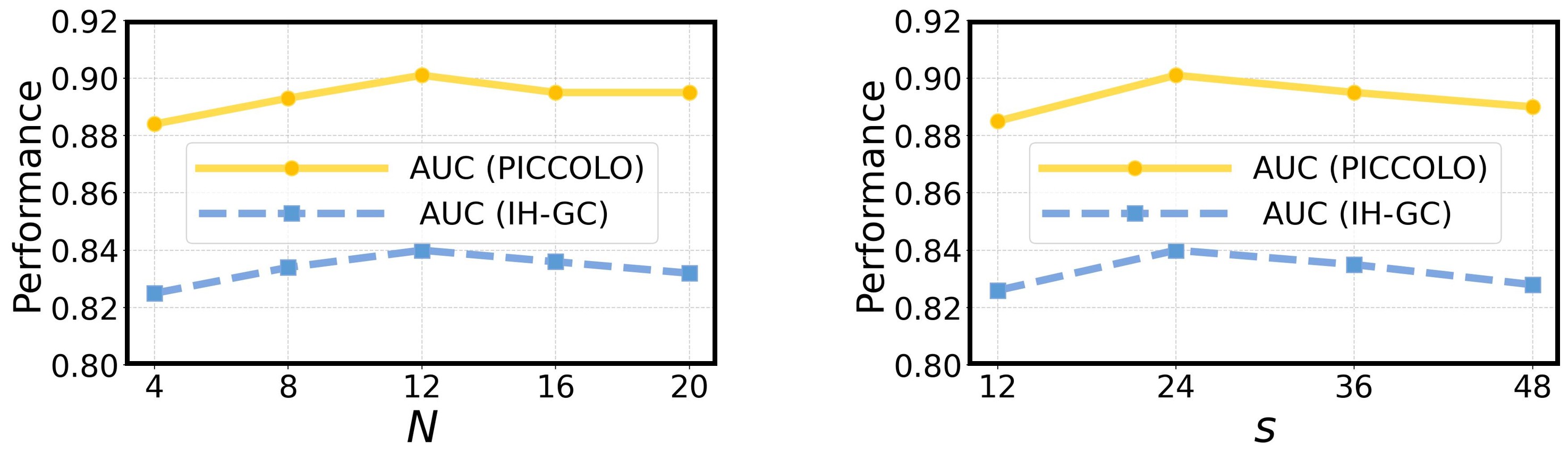}
    \caption{Ablation on two hyperparameters: $N$ and $s$.}
    \label{fig:ablation_hyperparameter}
\end{figure}

Figure~\ref{fig:ablation_hyperparameter} presents the results under different hyperparameter settings: the number of lesion queries in LR-QFormer $N$ and training batch size $s$, where the batch size $s$ is positively correlated with the group size of each class.
PaGKD using $N$=$12$ and $s$=$24$ shows the best performance, respectively, and we adopt this configuration as the default setting of PaGKD.

\subsection{Comparison with an Alternative Solution}
\begin{table}[t]
\begin{center}
\renewcommand\arraystretch{1}
\setlength{\tabcolsep}{5.2 pt}
\footnotesize
\begin{tabular}{l|ccc|ccc}
\hline

\hline
\multirow{2}{*}{Methods} &\multicolumn{3}{c|}{PICCOLO} &\multicolumn{3}{c}{IH-GC}\\
&ACC &F$1$ &AUC &ACC &F$1$ &AUC\\
\hline
\multicolumn{6}{l}{$\blacktriangleright$~\textit{PolypsAlign}}\\
$+$ CycleGAN &$70.2$ &$66.6$ &$78.8$ &$62.5$ &$66.6$ &$59.6$\\
$+$ SynDiff &$73.1$ &$69.0$ &$80.4$ &$69.6$ &$73.3$ &$67.7$\\
\hline
\multicolumn{6}{l}{$\blacktriangleright$~\textit{ADD}}\\
$+$ CycleGAN &$74.3$ &$71.3$ &$80.5$ &$66.1$ &$64.1$ &$71.5$\\
$+$ SynDiff &$74.5$ &$72.2$ &$81.7$ &$70.2$ &$66.9$ &$72.4$\\
\hline
PaGKD &$\mathbf{81.9}$ &$\mathbf{81.1}$ &$\mathbf{90.1}$ &$\mathbf{81.5}$ &$\mathbf{85.7}$ &$\mathbf{84.0}$\\
\hline

\hline
\end{tabular}
\caption{Comparisons with an alternative solution via the unpaired I2I translation method.} 
\label{table:comparsion_I2I}
\end{center}
\end{table}

Existing CDC methods like those in Table~\ref{table:compare_SOTAs} are designed for paired images, one possible workaround for unpaired data is to synthesize missing modalities via unpaired image-to-image (I2I) translation, enabling CDC methods to be effectively applied. 
To evaluate this alternative, we combine two representative I2I methods, CycleGAN~\cite{zhu2017unpaired} and SynDiff~\cite{ozbey2023unsupervised}, with the two best CDC methods excluding PaGKD in Table~\ref{table:compare_SOTAs}, \textit{i.e.}, PolypsAlign and ADD.
As shown in Table~\ref{table:comparsion_I2I}, our PaGKD remains superior to this alternative I2I-based solution.
These methods have not yet effectively enabled ADD and PolypsAlign to benefit from additional unpaired data, and in some metrics, even lag behind their counterparts trained solely on paired data.
The primary reason is that existing I2I methods focus on translating global imaging styles, such as imaging color, while neglecting modality translation of pathology-relevant content, such as vascular structure.
As a result, the synthesized images often lack pathological realism, reducing their utility for downstream classification.

\section{Conclusion and Future Work}
\label{sec:conclu}
In this paper, we propose PaGKD, which addresses the limitations of existing CDC methods that rely on paired data. Motivated by the semantic inconsistencies in unpaired data, we design two group-level distillation modules: GKD-Pro for learning modality-invariant semantic prototypes, and GKD-Den for capturing fine-grained structural features via group-level dense alignment. Our method can be applied to multi-modal learning scenarios without requiring instance-wise correspondence. Extensive experiments on four clinical datasets demonstrate the effectiveness and robustness of PaGKD in real-world GI classification, showcasing its strong clinical applicability. Nonetheless, PaGKD currently adopts a random group formation strategy; future work will explore an adaptive or data-driven grouping strategy.

\section{Acknowledgments}
\label{sec:ack}
This work was supported in part by National Natural Science Foundation of China (Grant No.62202189), research grants from Wuhan United Imaging Healthcare Surgical Technology Co., Ltd. Thanks to Professor Mei Liu’s team for providing the in-house dataset used in this work at the Department of Gastroenterology, Tongji Medical College, Huazhong University of Science and Technology.

\bibliography{reference}
\clearpage %

\appendix
\section{Appendix}
\label{sec:appendix}
\renewcommand{\thefigure}{A\arabic{figure}} 
\renewcommand{\thetable}{A\arabic{table}}
\setcounter{figure}{0}
\setcounter{table}{0}
\subsection{Grouping Details}
\label{appendix:grouping}
In the training, each batch includes all WLI/NBI groups per class, sized by class distribution. Groups are randomly reformed every 5 epochs (100 total) for diversity and stability, mitigating overfitting and unpaired cross-modal issues.

\subsection{Analysis of Receiver Operating Characteristic (ROC) curves}
In Table 2 of manuscript, we conducted a quantitative comparison of our PaGKD with eight methods, including four cross-modal independent classification (CIC) methods: SSL-CPCD~\cite{xu2024ssl}, SSL-WCE~\cite{guo2020semi}, FFCNet~\cite{wang2022ffcnet}, and DLGNet~\cite{wang2023dlgnet}, and four cross-modal distillation classification (CDC) methods: CPC-Trans~\cite{ma2022toward}, SAMD~\cite{shen2023auxiliary}, PolypsAlign~\cite{wang2021colorectal}, and ADD~\cite{hu2025holistic}.

\begin{figure}[h]
    \centering
    \includegraphics[width=0.93\linewidth]{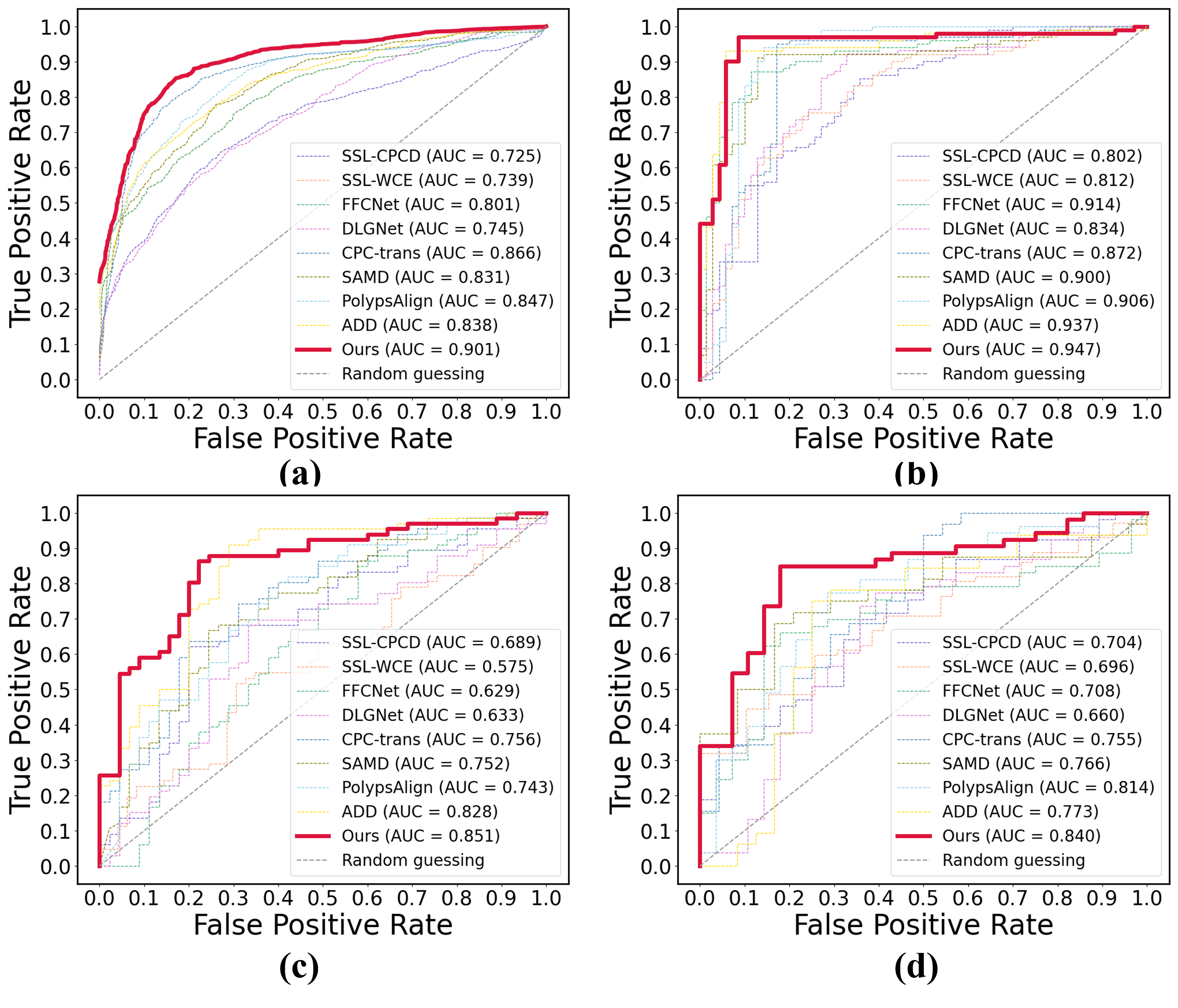}
    \caption{Receiver operating characteristic (ROC) curves on four datasets: (a) PICCOLO, (b) PolypSet (c) IH-Polyp (d) IH-GC.}
    \label{fig:roc}
\end{figure}

Furthermore, to more clearly demonstrate the performance differences among all comparison methods, we visualized the ROC curves of all methods on PICCOLO~\cite{sanchez2020piccolo} and IH-GC.
Among them, all the CDC methods except PaGKD are trained only on paired data because unpaired data can compromise their performance.
The ROC curve illustrates the trade-off between the true positive rate (TPR) and false positive rate (FPR), where a curve closer to the upper-left corner indicates better classification performance. As shown in Figure~\ref{fig:roc}, compared with other SOTA methods, our PaGKD consistently outperforms all baselines in this critical region, reflecting its ability to achieve high sensitivity while maintaining a low false alarm rate. Such performance is particularly important in clinical practice, where high TPR minimizes the risk of missed diagnoses, and low FPR reduces unnecessary treatments and additional diagnostic procedures. This highlights the clinical reliability and robustness of our approach under various operating conditions.
\label{appendix:ROC}

\subsection{Hyperparameters Configuration}
\label{appendix:hyperparameters}
To evaluate the impact of two hyperparameters $\tau_1$ and $\tau_2$ adopted in the semantic relation generation module, we conduct ablation studies on two datasets PICCOLO and IH-GC.
Specifically, we fix $\tau_2=0.7$ and adjust $\tau_1 \in \{0.1,0.2,0.3,0.4,0.5 \}$, and then fix $\tau_1=0.3$ and vary $\tau_2 \in \{0.5,0.6,0.7,0.8,0.9\}$.
\begin{table}[h]
\begin{center}
\renewcommand\arraystretch{1}
\setlength{\tabcolsep}{4.2 pt}
\footnotesize
\begin{tabular}{cccc|cccc}
\hline

\hline
$\tau_1$ &ACC &F$1$ &AUC &$\tau_2$&ACC &F$1$ &AUC\\
\hline
$0.1$&$0.787$&$0.778$&$0.863$&$0.5$&$0.788$&$0.765$&$0.855$\\
$0.2$&$0.788$&$0.784$&$0.878$&$0.6$&$0.796$&$0.783$&$0.889$\\
$\mathbf{0.3}$&$\mathbf{0.819}$&$\mathbf{0.811}$&$\mathbf{0.901}$&$\mathbf{0.7}$&$\mathbf{0.819}$&$\mathbf{0.811}$&$\mathbf{0.901}$\\
$0.4$&$0.809$&$0.801$&$0.892$&$0.8$&$0.788$&$0.771$&$0.879$\\
$0.5$&$0.789$&$0.773$&$0.864$&$0.9$&$0.787$&$0.764$&$0.849$\\
\hline

\hline
\end{tabular}
\caption{Ablation on two hyperparameters $\tau_1$ and $\tau_2$ on PICCOLO dataset.}
\label{table:ablation_piccolo}
\end{center}
\end{table}

\begin{table}[h]
\begin{center}
\renewcommand\arraystretch{1}
\setlength{\tabcolsep}{4.2 pt}
\footnotesize
\begin{tabular}{cccc|cccc}
\hline

\hline
$\tau_1$ &ACC &F$1$ &AUC &$\tau_2$&ACC &F$1$ &AUC\\
\hline
$0.1$&$0.774$&$0.793$&$0.798$&$0.5$&$0.786$&$0.800$&$0.808$\\
$0.2$&$0.788$&$0.821$&$0.823$&$0.6$&$0.786$&$0.828$&$0.825$\\
$\mathbf{0.3}$&$\mathbf{0.815}$&$\mathbf{0.857}$&$\mathbf{0.840}$&$\mathbf{0.7}$&$\mathbf{0.815}$&$\mathbf{0.857}$&$\mathbf{0.840}$\\
$0.4$&$0.786$&$0.815$&$0.816$&$0.8$&$0.788$&$0.813$&$0.817$\\
$0.5$&$0.776$&$0.795$&$0.804$&$0.9$&$0.764$&$0.782$&$0.793$\\
\hline

\hline
\end{tabular}
\caption{Ablation on two hyperparameters $\tau_1$ and $\tau_2$ on IH-GC dataset.}
\vspace{-0.5cm}
\label{table:ablation_EGC}
\end{center}
\end{table}
As shown in Table~\ref{table:ablation_piccolo} and Table~\ref{table:ablation_EGC}, the setting of $\tau_1=0.3$ and $\tau_2=0.7$ yields the best performance on both datasets, thus we employ this setting as the default setting.
This is attributed to the following: when $\tau_2$ is too large, the activated regions become overly restricted, leading to insufficient foreground coverage in semantic relations and hindering effective learning of lesion features. On the contrary, a small $\tau_2$ results in overly broad foreground regions, introducing noise and reducing the focus on critical lesions.
Similarly, an excessively large $\tau_1$ may cause potentially relevant lesion regions to be treated as background during distillation, thus confusing background and foreground semantics. In contrast, a very small $\tau_1$ limits the background to a small area, restricting the distillation of comprehensive background information and degrading background feature learning.

\subsection{Computational Complexity}
As shown in Table~\ref{table:ablation_computational_cost}, both the PaGKD-Pro and PaGKD-Den modules introduce only minimal computational overhead and a small number of additional parameters.
\begin{table}[h]
\begin{center}
\renewcommand\arraystretch{1}
\setlength{\tabcolsep}{4.5 pt}
\footnotesize
\begin{tabular}{lcc}
\hline

\hline
Component &Training Time (ms/iteration) &Parameters (M)\\
\hline
PaGKD-Pro &15.6 &0.07\\
PaGKD-Den &271.6 &0.16\\
\hline

\hline
\end{tabular}
\caption{Training time and number of parameters introduced by each component. The experiment is conducted on a single NVIDIA 4090 GPU with a batch size of $24$.}
\label{table:ablation_computational_cost}
\end{center}
\end{table}

\subsection{Extended Related Unpaired Multi-modal Works in Medical Image Analysis}
\subsubsection{Cross-modal Knowledge Transfer.}
The paradigm of enhancing model performance on a weaker modality by establishing \textit{knowledge transfer} from a dominant modality to a weaker one has been applied not only to the gastrointestinal endoscopy discussed in this paper (where NBI serves as the dominant modality and WLI as the weaker modality), but also to other medical scenarios: 
(1) Several studies~\cite{li2020towards,jiang2021unpaired} have improved segmentation performance on CT by transferring knowledge from MRI. These methods typically employ CycleGAN~\cite{zhu2017unpaired} to address the unpaired cross-modality issues. We also attempted to adopt this strategy in our study, as shown in Table~\ref{table:comparsion_I2I}, but the results were unsatisfactory. We attribute this to the fact that segmentation tasks only rely on image translators to achieve macroscopic-level transformations of organs or lesions, whereas gastrointestinal lesion classification depends on microscopic-level transformations involving vascular and mucosal structures. The naive existing image translation methods are insufficient to capture such fine-grained modality differences.
(2) Similarly, CoMoTo~\cite{alberb2024comoto} achieves knowledge transfer from mammography to digital breast tomosynthesis (DBT) to enhance breast cancer detection performance in DBT.
Since these approaches primarily focus on segmentation or detection tasks and depend on mask or bounding-box annotations during training, we do not directly compare our PaGKD with them in this work.

\subsubsection{Cross-modal Generalizability.}
There are also studies aimed at addressing how to enable models to achieve excellent multi-modal generalization performance on unpaired data.
\cite{dou2020unpaired} achieves outstanding performance with a highly compact architecture by constraining the divergence of prediction distributions across modalities.
MulModSeg~\cite{li2025mulmodseg} effectively integrates essential features from unpaired CT and MR images by introducing a modality-conditioned text embedding framework and an alternating training strategy.

\end{document}